\documentclass{article}

\usepackage[dblblindworkshop, final]{neurips_2025}

\usepackage[colorlinks=true, linkcolor=blue, urlcolor=blue, citecolor=blue]{hyperref}

\newcommand{\argmax}{\mathop{\mathrm{argmax}}\limits}

\usepackage[utf8]{inputenc} 
\usepackage[T1]{fontenc}    
\usepackage[dvipsnames]{xcolor}   

\usepackage{booktabs}       
\usepackage{amsfonts}       
\usepackage{nicefrac}       
\usepackage{microtype}      
\usepackage{amsmath}
\usepackage{graphicx} 
\usepackage{caption}  
\usepackage{float}    
\usepackage{subcaption} 
\usepackage{booktabs} 
\usepackage{enumitem}

\usepackage{algorithm}      
\usepackage{algpseudocode}  
\usepackage{algorithmicx}   

\usepackage{rotating}
\usepackage{tabularx}
\usepackage{booktabs}
\usepackage{adjustbox}
\usepackage{multirow}
\usepackage{dsfont}

\title{EQ-Negotiator: Dynamic Emotional Personas Empower Small Language Models for Edge-Deployable Negotiation}

\author{
Yunbo Long\textsuperscript{2,†} \hspace{0.5em}
Yuhan Liu\textsuperscript{1,†} \hspace{0.5em}
Liming Xu\textsuperscript{2} \hspace{0.5em}
Alexandra Brintrup\textsuperscript{2,3} \\
\textsuperscript{1}Rotman School of Management, University of Toronto, Canada \\
\textsuperscript{2}Department of Engineering, University of Cambridge, UK \\
\textsuperscript{3}The Alan Turing Institute, London, UK \\
yuhanlydia.liu@alumni.utoronto.ca, \{yl892, lx249, ab702\}@cam.ac.uk
}

\begin{document}

\maketitle

\begin{abstract}
The deployment of large language models (LLMs) in automated negotiation has set a high performance benchmark, but their computational cost and data privacy requirements render them unsuitable for many privacy-sensitive, on-device applications such as mobile assistants, embodied AI agents or private client interactions. While small language models (SLMs) offer a practical alternative, they suffer from a significant performance gap compared to LLMs in playing emotionally charged complex personas, especially for credit negotiation.
This paper introduces EQ-Negotiator, a novel framework that bridges this capability gap using emotional personas. Its core is a reasoning system that integrates game theory with a Hidden Markov Model (HMM) to learn and track debtor emotional states online, without pre-training. This allows EQ-Negotiator to equip SLMs with the strategic intelligence to counter manipulation while de-escalating conflict and upholding ethical standards.
Through extensive agent-to-agent simulations across diverse credit negotiation scenarios—including adversarial debtor strategies like cheating, threatening, and playing the victim—we show that a 7B parameter language model with EQ-Negotiator achieves better debt recovery and negotiation efficiency than baseline LLMs more than 10 times its size. 
This work advances persona modeling from descriptive character profiles to dynamic emotional architectures that operate within privacy constraints. Besides, this paper establishes that strategic emotional intelligence, not raw model scale, is the critical factor for success in automated negotiation, paving the way for effective, ethical, and privacy-preserving AI negotiators that can operate on the edge. Code is available at \url{https://github.com/Yunbo-max/EQ-Negotiator}.
\end{abstract}

\section{Introduction}\label{sec:introduction}







Large language models are increasingly deployed as negotiation agents representing diverse entities—from individuals and corporations to governmental bodies and autonomous systems \citep{zhu2025automated,long2025evoemoevolvedemotionalpolicies}. These agents access and process highly sensitive information during negotiations, including personal financial records, corporate strategies, medical data, and national security matters. However, the prevailing cloud-centric deployment paradigm creates fundamental vulnerabilities: sensitive data must be transmitted to external servers, exposing negotiations to privacy breaches and security risks \citep{he2024emerged,zhang2023sa}.

The edge-deployable negotiation paradigm we introduce addresses this critical limitation by enabling SLM agents to operate locally on end-user devices and institutional hardware. This approach has transformative implications across domains where privacy, latency, and autonomy are paramount. On mobile devices, edge-deployed agents can negotiate e-commerce transactions and social media interactions without exposing personal data. In autonomous systems, embodied AI—from industrial robots in warehouse management to wildfield search-and-rescue robots—need to conduct real-time environmental negotiations. This allows them to resolve conflicts over limited resources directly with other agents. In addition, major institutions including banks, hospitals, and corporations can deploy negotiation agents that access proprietary databases locally while interacting with external parties.

Despite these compelling applications, current SLM agents lack the enough emotional intelligence required for effective negotiations \citep{belcak2025small}. Their training on general emotional corpora creates critical weaknesses in adversarial contexts, where static emotional personas render them vulnerable to exploitation \citep{hu2024llm,orpek2024language}. This is particularly problematic in high-stakes domains like credit negotiations, where debtors strategically deploy aggression, feigned distress, or guilt-tripping to manipulate outcomes \citep{chen2024role}. Credit finance exemplifies these challenges, where effective negotiation requires balancing relationship preservation with accountability enforcement \citep{clempner2020shaping,prassa2020towards}. While affective computing shows promise in financial interactions \citep{yuasa2001negotiation, faure1990social}, current negotiation models lack strategic emotional depth \cite{schneider2024negotiating}. 
This gap is critical given emotional intelligence's recognized importance in financial negotiations \cite{hill2010emotionomics,marinkovic2015customers}.
The limitations manifest in three critical deficiencies: 1)Privacy-Emotion Tradeoff. Cloud-based emotional AI requires exposing sensitive negotiation data, while private edge solutions lack sophisticated emotional intelligence. 2) Manipulation Vulnerability. General emotional training leaves LLMs unable to distinguish genuine distress from strategic manipulation \citep{williams2024targeted}.
3) Contextual Rigidity. Unlike humans who adapt emotionally across interactions \citep{rawat2024little}, edge-deployable agents lack dynamic emotional adaptation.

\begin{figure*}[t] 
    \includegraphics[width=1\textwidth]{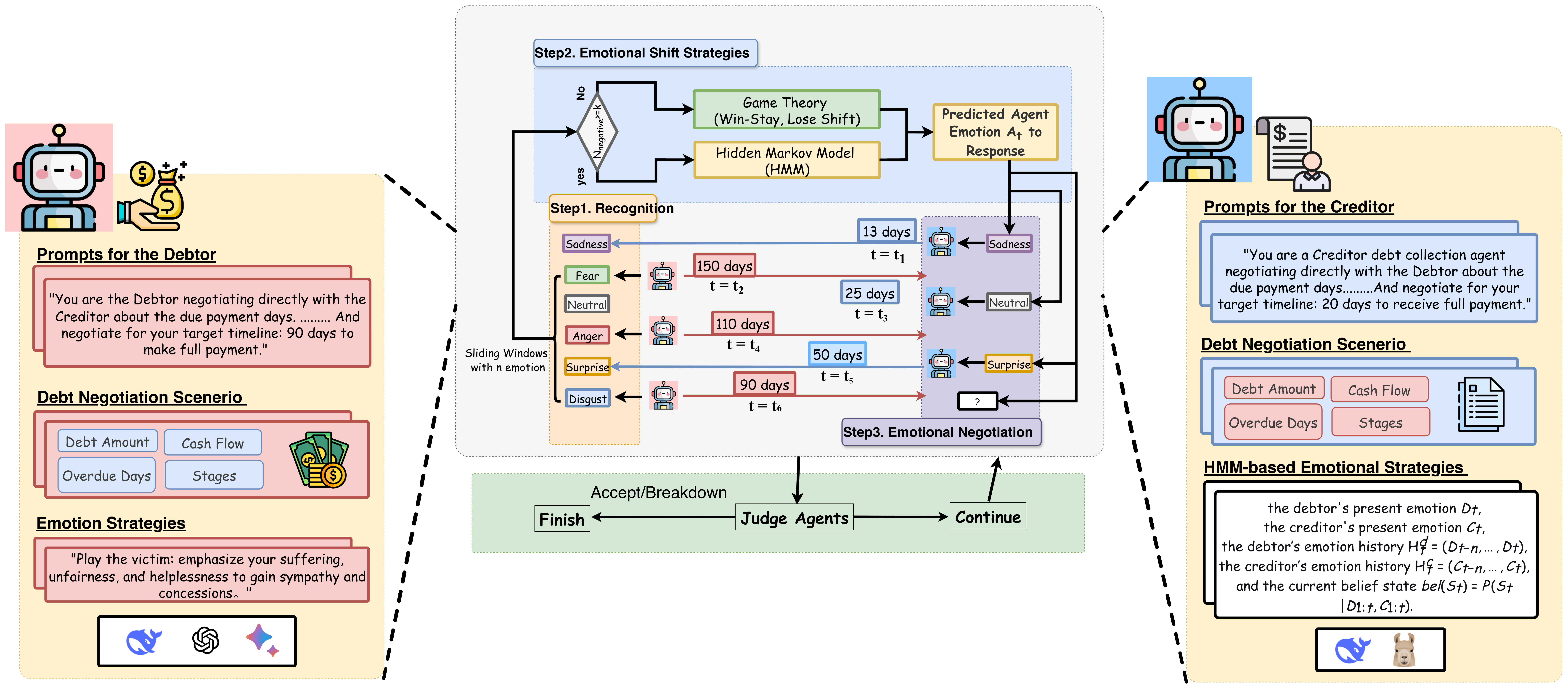}
    \caption{Pipeline of EQ-Negotiator: Integrating Emotional Intelligence for Automated Credit Negotiation} 
    \label{fig:workflow} 
\end{figure*}

Therefore, we present EQ-Negotiator, a novel emotional persona framework that enables effective negotiations with sophisticated emotional intelligence. Our approach transforms persona modeling from static profiles to dynamic emotional architectures, supporting edge-based applications from mobile commerce to institutional negotiations and autonomous system interactions.
This study addresses the limitations of current negotiation systems through EQ-Negotiator, which enables edge-based negotiation agents to:
\begin{itemize}[nosep]
\item Operate locally on edge devices using efficient 7B-parameter models, ensuring privacy-preserving financial negotiations without cloud dependency.
\item Learn negotiation strategy online by employing Hidden Markov Models (HMMs) to infer debtor emotional states and interaction patterns directly from the ongoing dialogue, eliminating the need for pre-trained emotional models.
\item Execute dynamic emotional policies that adapt in real-time, strategically shifting between emotional personas to de-escalate conflict and effectively counter adversarial tactics.
\end{itemize}

By grounding persona expression in computational emotion dynamics rather than static personality traits, EQ-Negotiator transforms edge-deployable SLMs agents from vulnerable empathy simulators into strategically robust negotiation partners capable of navigating extreme emotional scenarios.

\section{Related Work}\label{sec:related_work}

\subsection{Edge-Deployable AI Agents in Negotiation}

Autonomous LLM agents have been increasingly applied to role-playing scenarios such as card games like Avalon, financial trading, and credit collection, where they simulate negotiating parties \citep{light2023avalonbench}. However, current research typically assumes cloud-based LLM agents can directly access sensitive information from banks, hospitals, or personal devices during negotiations, overlooking critical data privacy and security concerns inherent in prompt-based information transmission strategies \citep{he2024emerged}. Furthermore, existing methods heavily depend on network connectivity, suffering from latency issues that can degrade user experience \citep{belcak2025small}.
Particularly important are scenarios affected by geopolitical constraints and corporate policies, where LLM API access may be restricted or disabled. In remote areas with limited connectivity, or in embodied AI systems where robot swarms must negotiate with humans in real-time, the dependency on cloud services becomes a fundamental limitation. These challenges underscore the urgent need for offline, lightweight agents deployable on edge devices, enabling robust and private negotiations without external dependencies while ensuring data sovereignty and operational resilience in diverse environmental conditions.

\subsection{Small Language Models in Negotiation}

The paradigm of language models is broadly divided into Large Language Models (LLMs) and Small Language Models (SLMs), with the latter typically defined as models with 7 billion parameters or fewer \citep{belcak2025small}. While LLMs have demonstrated remarkable, emergent abilities in general tasks, their massive scale induces critical limitations, including high computational demands, privacy concerns from cloud dependency, and unsuitability for real-time, edge-device applications \citep{orpek2024language}. Consequently, SLMs have gained prominence for their low latency, cost-effectiveness, and ease of customization. However, a well-documented performance gap persists, primarily attributed to the scaling laws; SLMs inherently lack the extensive world knowledge and nuanced reasoning capabilities of LLMs \citep{lu2024small}. Existing research has extensively documented the shortcomings of SLMs in mathematical and commonsense reasoning. 
However, a critical gap remains in assessing their capacity for emotional intelligence—specifically, the ability to adopt an emotional persona, infer others' emotional states, and adapt strategies in real-time during socio-emotional interactions such as negotiation.

\subsection{Emotional Intelligence in Credit Negotiation}

While LLMs show promise in automating debt negotiation, few studies have examined their emotional strategies in this context. \citet{schneider2024negotiating} applied LLMs to price negotiations with humans but overlooked the role of emotional dynamics. Similarly, \citet{wang2025debt} found that LLMs tend to over-concede compared to human negotiators and proposed a Multi-Agent approach to improve decision rationality, yet they did not account for the function of emotions in negotiation. Typically, LLM-based agents mimic empathy by recognizing patterns in their training data rather than employing strategic emotional reasoning. Without genuine affective understanding, they struggle to adjust their tone and negotiation strategy based on a debtor's emotional state. For instances, if a debtor gets angry, the agent may escalate tension. And if the debtor sound desperate, the agent might concede unfairly. Without emotional intelligence, negotiation agents are not only inefficient but strategically vulnerable, making them easy targets for manipulation and exploitation by sophisticated, emotionally-aware agents.

\section{EQ-Negotiator}

As shown in \autoref{fig:workflow}, EQ-Negotiator is an emotion-aware negotiation framework designed to dynamically adapt emotional tones during credit negotiations, enabling deployment on edge devices. The system operates within a dual-agent simulation framework, where a \textit{creditor agent} interacts with a \textit{debtor agent} in realistic credit scenarios. Rather than relying on fine-tuned pre-trained language models, our approach directly leverages the inherent capabilities of language models to recognize debtor emotion in real time from dialogue context. A key component of the framework is its emotional memory mechanism, which records historical emotional interactions and serves as a contextual buffer for tracking debtor sentiment shifts across turns. Based on the observed and remembered emotional states, the system employs game theory and hidden Markov models to dynamically adjust the creditor's own emotional tone throughout the negotiation. The HMM maintains an internal hidden state representing the creditor's strategic mode, which is used to make optimal decisions about which emotional expression to employ at each turn. To enable fully automated and reproducible evaluation, we implement a multi-agent assessment system in which a separate \textit{negotiation judge agent} determines the success or breakdown of each negotiation round. Details see Algorithm \autoref{alg:eq-negotiator}.

\subsection{In-Context Emotion Recognition}

Our framework performs emotion recognition through in-context learning, eliminating fine-tuning requirements while maintaining edge deployment efficiency. For each debtor utterance, the system constructs prompts containing: (1) definitions of seven emotional states, (2) conversational examples, and (3) current dialogue context. The complete template is detailed in Appendix \ref{app:prompts}. Besides, the system maintains dual emotion histories: $\mathcal{H}_t^d = (D{t-n}, \dots, D_t)$ for debtor emotions and $\mathcal{H}_t^c = (C{t-n}, \dots, C_t)$ for creditor emotions, where $D_i, C_i \in \mathcal{E} = {\text{Joy}, \text{Sadness}, \text{Anger}, \text{Fear}, \text{Disgust}, \text{Surprise}, \text{Neutral}}$. These sequences enable pattern detection for HMM-based strategies.

\subsection{Emotional Shift Strategies}

We combine game theory and Hidden Markov Models for dynamic emotion adaptation. 

\paragraph{Win-Stay, Lose-Shift}
Instead of using a pure Tit-for-Tat (TFT) strategy, we propose a novel Win-Stay, Lose-Shift (WSLS) with emotional weighting (Table~\ref{tab:complete_payoff}). For positive debtor emotions (Joy, Neutral, Surprise), the creditor maintains cooperation; for negative exchanges (Anger, Disgust, Fear), it shifts to cautious responses. This avoids the escalation risks of pure Tit-for-Tat while providing necessary resistance in credit negotiations where trust and risk management are critical.
The emotional dynamics form a repeated game where the creditor employs:
\begin{equation}
f_{\text{Payoff}}(d) = \argmax_{e \in \mathcal{E}} \pi(d, e)_2
\end{equation}
with $\pi(d, e) = (\pi_1(d,e), \pi_2(d,e))$ representing debtor and creditor payoffs respectively. See details of this algorithm in Appendix \ref{app:wsls}.

\paragraph{Hidden Markov Model for Emotional Strategies}
Our system employs a \textit{Hidden Markov Model} (HMM) to model the temporal dynamics of emotional interactions between creditor and debtor. The hidden states represent the creditor's strategic mode $S_t$ $\in \{\text{Cooperative Mode},\, \text{Confrontational Mode},\, \text{Distressed Mode},\, \text{Strategic Mode}\}$,
 while the observable variables are the emotional exchange $(D_t, C_t)$ at each turn, where $D_t$ is the debtor's emotion and $C_t$ is the creditor's expressed emotion. 
 The details of those prompts are in Appendix \ref{app:prompts}.
The HMM captures the sequential dependency through:
\begin{itemize}
\item \textbf{Transition Probability}: $P(S_{t+1} \mid S_t, D_t, C_t)$ - how the emotional context evolves given the current interaction
\item \textbf{Emission Probability}: $P(D_t, C_t \mid S_t)$ - how hidden emotional context manifests as observable emotion pairs
\end{itemize}

The system maintains a belief state $bel(S_t) = P(S_t \mid D_{1:t}, C_{1:t})$ updated via Bayesian filtering:
\begin{equation}
bel(S_t) = \eta \cdot P(D_t, C_t \mid S_t) \cdot \sum_{S_{t-1}} P(S_t \mid S_{t-1}, D_{t-1}, C_{t-1}) \cdot bel(S_{t-1})
\end{equation}
where the normalization constant $\eta$ ensures the belief state sums to 1 over all possible states:
\begin{equation}
\eta = \frac{1}{\sum_{S_t} \left[ P(C_t \mid S_t) \cdot \sum_{S_{t-1}} P(S_t \mid S_{t-1}, D_{t-1}, C_{t-1}) \cdot bel(S_{t-1}) \right]}
\end{equation}
This normalization constant is calculated as the reciprocal of the sum of all unnormalized belief values across all possible states $S_t \in \mathcal{S}$.
When debtors exhibit $\geq k$ negative emotions within an n-turn window, the HMM activates to predict optimal creditor responses. Let $\mathcal{H}t^d = (D{t-n}, \dots, D_t)$ be the debtor's emotion history and $\mathcal{H}t^c = (C{t-n}, \dots, C_t)$ be the creditor's emotion history. The policy selection is:
\begin{equation}
C_{t+1} =
\begin{cases}
f_{\mathrm{HMM}}\bigl(D_t,\, C_t,\, \mathcal{H}_t^d,\, \mathcal{H}_t^c\bigr)
&\text{if }\; \sum_{i=t-n+1}^{t} \mathds{1}\{D_i \in \mathcal{E}_{\mathrm{neg}}\} \ge k,\\[6pt]
f_{\mathrm{Payoff}}(D_t) &\text{otherwise.}
\end{cases}
\end{equation},
where $\mathcal{E}_{\text{neg}} = {\text{Sadness}, \text{Anger}, \text{Fear}, \text{Disgust}}$.
The HMM-based decision policy predicts the next optimal creditor emotion by maximizing expected utility over possible emotional trajectories:
\begin{equation}
\footnotesize
f_{\text{HMM}}(D_t, C_t, \mathcal{H}^d_t, \mathcal{H}^c_t) = \argmax_{e \in \mathcal{E}} \sum_{S_{t+1}} \Big[
P(S_{t+1} \mid \mathcal{H}^d_t, \mathcal{H}^c_t) \cdot P(D_{t+1} \mid S_{t+1}, C_t = e) \cdot w(e, D_t, S_{t+1})
\Big]
\end{equation}

The hidden state prediction incorporates the complete interaction history:
\begin{equation}
P(S_{t+1} \mid \mathcal{H}^d_t, \mathcal{H}^c_t) = \sum_{S_t} P(S_{t+1} \mid S_t, D_t, C_t) \cdot P(S_t \mid D_{1:t}, C_{1:t})
\end{equation}

\noindent\textbf{Parameter Learning via Maximum Likelihood:}
The HMM parameters are learned from historical emotion sequences:
\begin{equation}
\theta^* = \argmax_{\theta} \sum_{m=1}^M \log P(D_{1:T}^{(m)}, C_{1:T}^{(m)} \mid \theta)
\end{equation}
where $\theta = {P(S_1), P(S_{t+1}|S_t, D_t, C_t), P(D_t, C_t|S_t)}$ and $(D_{1:T}^{(m)}, C_{1:T}^{(m)})$ are observed emotion interaction sequences.
This approach enables the creditor to anticipate emotional dynamics and select responses that strategically influence the interaction trajectory, balancing immediate payoff optimization with long-term emotional alignment.

\subsection{Multi-Agent Negotiation Simulation}

The complete EQ-Negotiator framework operates through an automated multi-agent simulation system, as formalized in Algorithm \autoref{alg:eq-negotiator}. The simulation involves three specialized agents: a \textit{debtor agent} ($\mathcal{M}_{debtor}$) that generates client responses, a \textit{negotiator agent} ($\mathcal{M}_{negotiator}$) that employs emotional intelligence strategies, and a \textit{judge agent} that evaluates negotiation outcomes. Each negotiation cycle begins with emotion recognition of the client's utterance, followed by strategic emotion selection using either the WSLS payoff maximization approach or the HMM-based Bayesian filtering method when negative emotion persistence is detected. The selected emotional state guides the generation of the negotiator's response through emotion-conditioned prompting. The judge agent continuously monitors the dialogue for agreement signals or breakdown conditions, ensuring objective evaluation of negotiation success across diverse scenarios.

\begin{algorithm}
\caption{Multi-Agent Credit Negotiation System}
\label{alg:eq-negotiator}
\begin{algorithmic}[1]
\State \textbf{Agents:} Creditor $\mathcal{M}_{creditor}$, Debtor $\mathcal{M}_{debtor}$, Judge $\mathcal{M}_{judge}$
\State \textbf{Data:} Dialogue history $H$, Emotion histories $\mathcal{H}^d, \mathcal{H}^c$

\Procedure{MainNegotiation}{}
    \State Initialize $H_0 \gets \emptyset$, $C_0 \gets \text{Neutral}$, $\mathcal{H}^d_0 \gets \emptyset$, $\mathcal{H}^c_0 \gets \emptyset$
    
    \For{$t = 0$ to $T_{max}$}
        \State $msg_{debtor} \gets \mathcal{M}_{debtor}(H_t)$, $D_t \gets \text{EmotionRecognition}(msg_{debtor})$
        \State $\mathcal{H}^d_t \gets \text{UpdateHistory}(\mathcal{H}^d_{t-1}, D_t)$
        
        \If{$\sum_{i=t-n+1}^{t} \mathds{1}[D_i \in \mathcal{E}_{neg}] \geq k$}
            \State $C_{t+1} \gets \argmax_{e \in \mathcal{E}} \sum_{S_{t+1}} P(S_{t+1} \mid \mathcal{H}^d_t, \mathcal{H}^c_t) \cdot P(D_{t+1} \mid S_{t+1}, C_t = e) \cdot w(e, D_t, S_{t+1})$
        \Else
            \State $C_{t+1} \gets \argmax_{e \in \mathcal{E}} \pi(D_t, e)_2$
        \EndIf
        
        \State $msg_{creditor} \gets \mathcal{M}_{creditor}(\text{EmotionPrompt}(C_{t+1}, H_t))$
        \State $H_{t+1} \gets H_t \cup \{(D_t, msg_{debtor}, C_{t+1}, msg_{creditor})\}$
        \State $\mathcal{H}^c_t \gets \text{UpdateHistory}(\mathcal{H}^c_{t-1}, C_{t+1})$
        
        \If{$\mathcal{M}_{judge}.\text{AgreementReached}()$ \textbf{or} $\mathcal{M}_{judge}.\text{Breakdown}()$} \textbf{break}
        \EndIf
    \EndFor
    \State \textbf{return} $H_{final}$, $\text{outcome}$
\EndProcedure
\end{algorithmic}
\end{algorithm}

\section{Experimental Settings}
\label{app:settings}

\paragraph{Credit Negotiation Dataset.}
\label{data:debt}

We employ the Credit Recovery Assessment Dataset (CRAD)\citep{long2025emodebtbayesianoptimizedemotionalintelligence}, which comprises 100 carefully constructed commercial delinquency scenarios developed specifically for debt recovery research. This comprehensive collection spans diverse financial situations with loan amounts ranging from \$20,688 to \$49,775 and delinquency periods extending from one month to nearly a year. The dataset encompasses multiple business sectors including manufacturing, retail, and technology, with varied credit arrangements such as working capital loans and commercial mortgages. Each scenario includes detailed contextual information about collateral types, recovery stages, and cash flow conditions, providing a realistic foundation for evaluating negotiation strategies across different financial distress situations. 

\paragraph{Experimental Configuration.}

Our experimental framework evaluates EQ-Negotiator through comprehensive multi-agent simulations designed to assess emotional intelligence in credit negotiation scenarios. The system configuration employs a dual-agent architecture where creditor agents equipped with our EQ-Negotitaor or orginal settings without any emotional guidance with debtor agents exhibiting diverse emotional or personas patterns.
We establish a rigorous evaluation protocol with the following parameters: a)HMM Activation Threshold: We set the HMM activation threshold at $k=4$ negative emotions within $n=5$ turns, an empirically effective criterion for identifying negotiations that benefit from emotional adaptation strategies. b) Debtor Personas: Four distinct debtor strategies are implemented: (1) Vanilla (baseline behavior without emotional guidance), (2) Fixed negative Emotion (persistent emotional states including anger, sadness, fear, and disgust). (3)Four distinct adversarial strategies:  Intimidation (threatening and aggressive tactics), Cheating Persona (systematic deceptive behavior) and  and Stonewalling Tactics (refusal to engage and deliberate response delays). c)Agent models: We evaluate across model scales including SLMs (DeepSeek-7B, Llama-7B) and LLMs (GPT-4o-mini, GPT-5-mini) to assess scalability and generalization, where DeepSeek-7B refers to DeepSeek-LLM-7B-Chat and Llama-7B refers to Llama-2-7b-chat-hf.
All experiments employ GPT-4o-mini as the fixed debtor agent, while varying creditor agent models.
Besides, our framework employs three core matrices that define the emotional interaction dynamics. The transition matrix (Table~\ref{tab:hmm_transition}) initializes HMM state transition probabilities with values reflecting emotional persistence in negotiation dialogues. The emission matrix (Table~\ref{tab:hmm_emission}) models the probabilistic relationship between agent and client emotions, capturing emotional mirroring effects commonly observed in human interactions. The payoff matrix (Table~\ref{tab:complete_payoff}) encodes strategic utilities for emotional interactions, where each entry $(x,y)$ represents the (client payoff, agent payoff) pair. All matrices undergo continuous refinement through online learning during negotiations to adapt to individual debtor behavior patterns. See the details in Appendix \ref{app:Psychological Foundations}.

\paragraph{Evaluation Metrics.}

We employ three key metrics to evaluate negotiation performance: success rate( measuring the proportion of successful negotiation agreements), debt collection multiples (calculated as final payment days divided by creditor's initial proposed days), and negotiation speed (measured in total dialogue rounds). For each metric except the success rate, we report the mean along with 95\% confidence intervals computed using a t-distribution approach that ensures non-negative bounds for continuous positive-valued metrics. All results are aggregated across 50 different scenarios with consistent random seeds for each model configuration to ensure statistical reliability and reproducibility.

To quantitatively evaluate ethical implications, we employ GPT-5 as an impartial evaluator to compute four metrics per scenario: 1) Manipulative Language (counting surprise expressions and pressure tactics), 2)False Empathy (identifying empathy statements without meaningful concessions), 3)Rigid Negotiation (detecting repeated offers or minimal changes where $|\Delta \text{days}| < 2$), and 4)Psychological Pressure (capturing gaslighting language and coercive tactics). Each metric is formalized as $X_m = \frac{1}{N}\sum_{i=1}^{N} \sum_{j=1}^{T_i} \mathds{1}(\text{condition}_{ij})$, representing the average count per scenario, where $\mathds{1}(\cdot)$ is the indicator function, $i$ indexes scenarios, $j$ indexes dialogue turns within each scenario, and $T_i$ denotes the total number of turns in scenario $i$. The ethical analysis was conducted five times, each on distinct negotiation results across all fixed debtor emotion settings, and we report the mean metric values.

\section{Experimental Results}

\subsection{Comparison of EQ-Negotiator with SLMs and LLMs in Vanilla Baseline}

\begin{table}[h]
\centering
\caption{Performance Evaluation of EQ-Negotiator (95\% Confidence Intervals) under the Vanilla settings of Debtors.}
\label{tab:comprehensive_results_1}
\scalebox{0.9}{%
\begin{tabular}{@{}lcccccc@{}}
\toprule
{\textbf{Creditor}} & \multicolumn{2}{c}{\textbf{Success Rate (\%)} $\uparrow$} & \multicolumn{2}{c}{\textbf{Debt Collection Multiples (x)} $\downarrow$} & \multicolumn{2}{c}{\textbf{Negotiation Speed} $\downarrow$} \\
\cmidrule(lr){2-3} \cmidrule(lr){4-5} \cmidrule(lr){6-7}
 & \textit{Original} & \textit{EQ-Neg} & \textit{Original} & \textit{EQ-Neg} & \textit{Original} & \textit{EQ-Neg} \\
\midrule
\textbf{GPT-5-mini} & 
85.0 & 90.0 & 
3.1 [2.2-4.1] & 3.7 [0.9-6.5] & 
7.5 [3.9-15.8] & 8.1 [3.2-14.2] \\

\textbf{GPT-4o-mini} & 
95  & 100  & 
3.6 [1.5-5.7] & 2.4 [2.1-2.7] & 
5.8 [1.8-9.8] & 3.7 [2.8-7.6] \\

\textbf{Deepseek-7b} & 
75.0  & 80  & 
5.5 [0.7-9.5] & 3.1 [1.6-6.5] & 
13.4 [1.7-16.4] & 9.4 [1.4-12.8] \\

\textbf{Llama-7b} & 
40.0  & 70  & 
6.1 [3.4-11.3] & 3.4 [1.5-7.4] & 
18.1 [5.7-21.4] & 14.8 [1.1-15.6] \\
\bottomrule
\end{tabular}%
}
\end{table}

The experimental results in Table \ref{tab:comprehensive_results_1} compellingly demonstrate that the EQ-Negotiator framework acts as a powerful emotional engine, enhancing the performance of all creditor models, but with a transformative impact particularly on Small Language Models (SLMs). While both the LLM (GPT-4o-mini) and SLMs (DeepSeek-7B, Llama-7B) saw improvements, the gains for the latter were disproportionately dramatic: Llama-7B's success rate surged from a baseline of 40\% to 70\%, and DeepSeek-7B, when equipped with the EQ-Negotiator,  reduced the performance gap compared to the original GPT-4o-mini in key metrics. This indicates that EQ-Negotiator can be effective for all negotitaion models while helpful to compensate for the inherent scale and parameter disadvantages of SLMs. Furthermore, the standout performance of the GPT-4o-mini creditor against a debtor of the same model—achieving a perfect 100\% success rate, the lowest debt multiple (2.4x), and the fastest negotiation speed (3.7 rounds)—suggests a unique "model homophily" effect. This indicates that negotiations between identical model architectures reach consensus more effectively than heterogeneous model pairings, potentially due to shared reasoning patterns and communication protocols that facilitate mutual understanding.

\subsection{Performance Against Emotionally Debtors}

\begin{table}[h]
\centering
\caption{Performance Evaluation of EQ-Negotiator (95\% Confidence Intervals) under Debtor Settings with negative emotion}
\label{tab:comprehensive_results_2}
\scalebox{0.8}{%
\begin{tabular}{@{}lcccccccc@{}}
\toprule
{\textbf{Model}} & {\textbf{Debtor}} & \multicolumn{2}{c}{\textbf{Success Rate (\%)} $\uparrow$} & \multicolumn{2}{c}{\textbf{Debt Collection Multiples (x)}} & \multicolumn{2}{c}{\textbf{Negotiation Speed}} \\
\cmidrule(lr){3-4} \cmidrule(lr){5-6} \cmidrule(lr){7-8}
 & & \textit{Original} & \textit{EQ-Neg} & \textit{Original} & \textit{EQ-Neg} & \textit{Original} & \textit{EQ-Neg} \\
\midrule
\multirow{4}{*}{\textbf{Deepseek-7B}} 
& \textbf{Sadness} & 
50.0 &75.0  & 
2.0 [1.3-2.8] & 1.9 [1.2-2.6] & 
8.5 [2.8-12.2] &7.6 [2.6-11.0] \\

& \textbf{Fear} & 
70.0  & 80.0 & 
3.7 [1.0-7.9] & 1.7 [1.2-2.9] & 
7.2 [1.5-10.1] & 4.4 [2.1-6.2] \\

& \textbf{Anger} & 
75.0  & 75.0 & 
3.1 [2.0-3.5] & 2.2 [1.9-2.4] & 
9.2 [2.9-10.1] & 7.9 [1.9-8.9] \\

& \textbf{Disgust} & 
40.0 & 60.0  & 
8.2 [1.2-10.8] & 3.7 [1.5-7.5] & 
15.2 [8.9-20.1] & 8.9 [0.5-14.4] \\

\midrule
\multirow{4}{*}{\textbf{GPT-4o-mini}}
& \textbf{Sadness} & 
80.0  & 100.0 & 
4.1 [1.9-6.2] & 4.1 [1.67-7.5] & 
6.1 [2.7-11.5] & 5.7 [2.2-9.1] \\

& \textbf{Fear} & 
100.0  & 95.0  & 
2.4 [1.6-4.1] & 3.1 [1.1-5.2] & 
4.1 [2.1-5.81] & 5.7 [1.7-7.6] \\

& \textbf{Anger} & 
100.0  & 100.0  & 
5.5 [2.5-8.4] & 4.7 [2.1-7.4] & 
10.0 [5.5-14.5] & 8.5 [2.2-14.8] \\

& \textbf{Disgust} & 
85.0 & 85.0 & 
3.5 [2.0-6.5] & 2.9 [1.9-3.8] & 
6.6 [2.4-13.0] & 4.7 [2.3-8.4] \\
\bottomrule
\end{tabular}%
}
\end{table}

The reulst in Table \ref{tab:comprehensive_results_2} reveals that the EQ-Negotiator consistently enhances performance for both LLMs and SLMs when dealing with emotionally charged debtors, but it does so in characteristically different ways. For the SLM (DeepSeek-7B), the EQ framework acts as a crucial stabilizer and optimizer, delivering dramatic improvements in the most challenging scenarios. This is most evident with a Disgusted debtor, where the success rate jumps from 40\% to 60\%, the Debt Collection Multiple is more than halved (from 8.2x to 3.7x), and negotiation speed is nearly doubled. Similarly, for a Sad debtor, the success rate surges by 25 percentage points. This demonstrates that the EQ-Negotiator provides the smaller model with the strategic empathy and tactical flexibility it inherently lacks, allowing it to de-escalate tension and secure better terms.

The intriguing observation that the optimized SLM achieves smaller debt collection days (a lower Multiple) despite longer negotiation rounds compared to the LLM, showing a fundamental difference in negotiation strategy. The SLM, guided by the EQ-Negotiator, engages in a more patient, rapport-building dialogue. It spends additional rounds validating the debtor's emotion and building trust, which then allows it to secure a more favorable, faster-repayment agreement. In contrast, the LLM (GPT-4o-mini), even with the EQ-Negotiator, often relies on its inherent logics to navigate emotions more efficiently, sometimes closing deals faster but conceding slightly more on the repayment timeline. This is clear in the Fear condition, where the EQ-powered DeepSeek-7B reduces the Debt Multiple to 1.7x in 4.4 rounds, while the EQ-powered GPT-4o-mini achieves a 3.1x Multiple in 5.7 rounds. Ultimately, the SLM, when equipped with expert emotional intelligence, adopts a "slow and steady wins the race" approach, leveraging prolonged, empathetic engagement to achieve superior financial outcomes, even against a more powerful base model.

\subsection{Performance in Countering Adversarial Debtor Strategies}

\begin{figure*}[htp!]
\centering
\begin{subfigure}{0.49\textwidth}
\centering
\includegraphics[width=\linewidth]{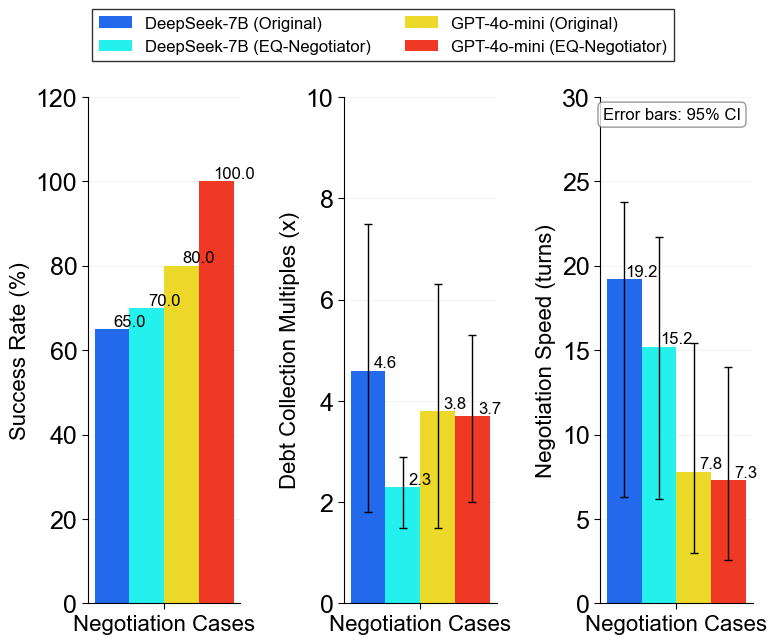}
\caption{Debtor with Cheating Tactics}
\label{fig:matrix_cheating}
\end{subfigure}
\hfill
\begin{subfigure}{0.49\textwidth}
\centering
\includegraphics[width=\linewidth]{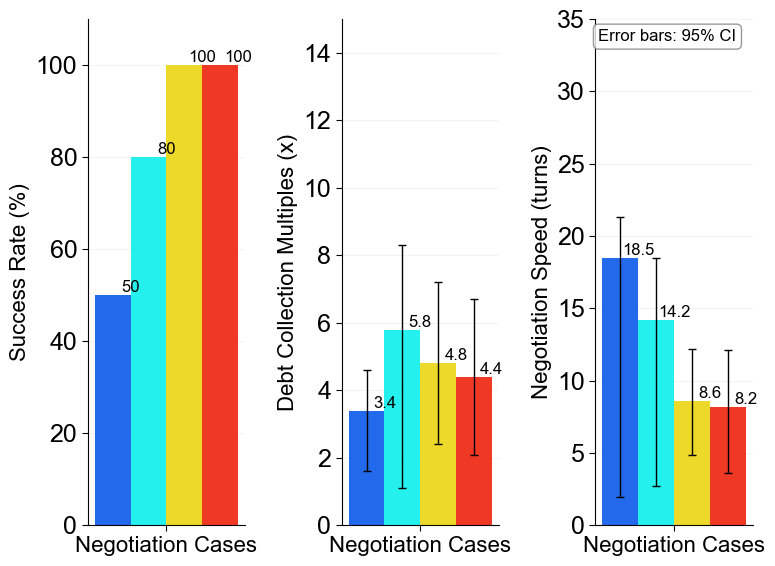}
\caption{Debtor Playing Victim}
\label{fig:matrix_victim}
\end{subfigure}

\vspace{0.5cm}

\begin{subfigure}{0.49\textwidth}
\centering
\includegraphics[width=\linewidth]{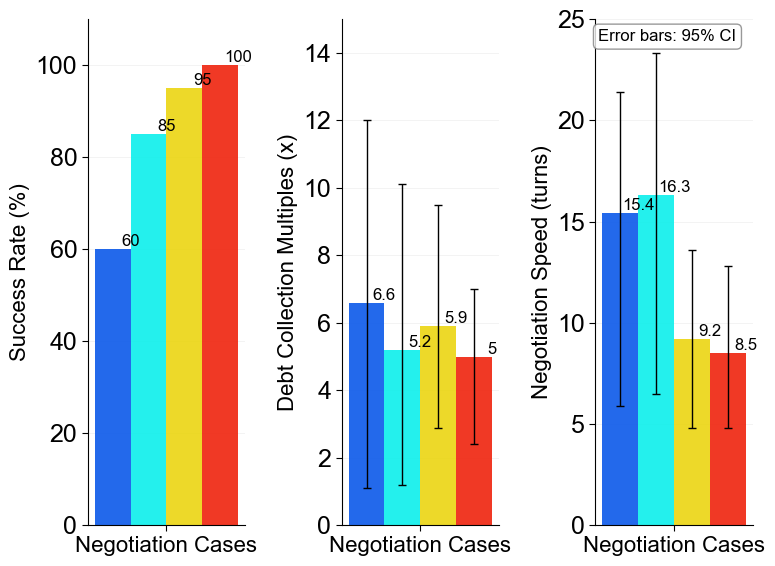}
\caption{Debtor with Threatening Tactics}
\label{fig:matrix_threatening}
\end{subfigure}
\hfill
\begin{subfigure}{0.49\textwidth}
\centering
\includegraphics[width=\linewidth]{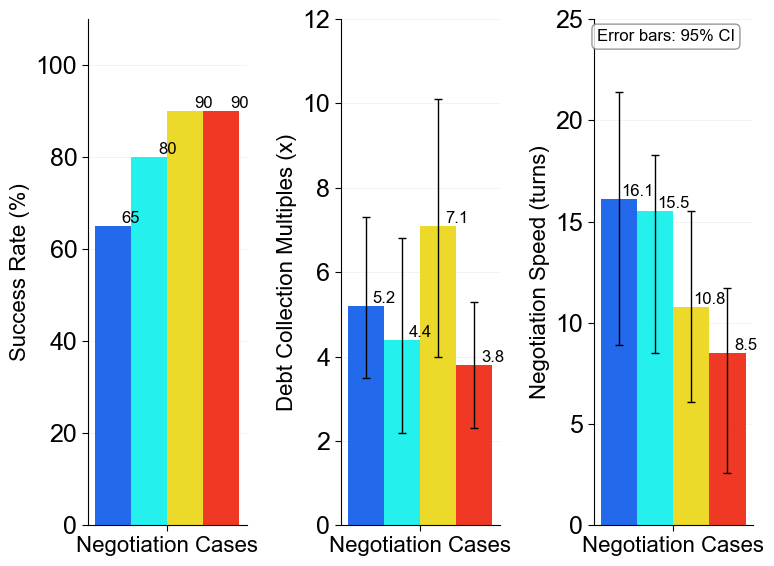}
\caption{Debtor with Stonewalling Tactics}
\label{fig:matrix_stonewalling}
\end{subfigure}

\caption{Debt collection performance of LLMs and EQ-enhanced SLMs when countering adversarial debtor strategies.}
\label{fig:optimized_matrices}
\end{figure*}

We introduce a more challenging negotiation environment where debtors employ explicit manipulation tactics rather than general emotional states. In this adversarial context, as shown in Figure~\ref{fig:optimized_matrices}, LLMs naturally excel due to their broader training and inherent reasoning capabilities, allowing them to better detect and counter sophisticated manipulation strategies like cheating, victim-playing, and threats.
However, in adversarial scenarios involving Threatening and Stonewalling tactics, EQ-enhanced SLMs achieve superior debt collection multiples compared to vanilla LLMs, albeit with slightly longer negotiation rounds. This demonstrates that while raw model scale provides inherent advantages, strategic emotional intelligence can effectively bridge this capability gap. 

The EQ-Negotiator essentially provides SLMs with the guidance for emotional dynamics that LLMs might infer naturally, enabling smaller models to achieve competitive negotiation outcomes through guided emotional adaptation. Notably, GPT-4o-mini with EQ-Negotiator consistently outperforms its vanilla counterpart across all metrics, confirming that emotional intelligence augmentation benefits models regardless of their scale in challenging negotiation scenarios.

\subsection{Ethical Analysis of EQ-Negotiator}

\begin{table}[h]
\centering
\caption{Ethical Behavior Comparison Between AI Negotiation Agents}
\label{tab:ethical_metrics}
\begin{tabular}{@{}lcccc@{}}
\toprule
\textbf{Ethical Metric} & \textbf{GPT-4o-mini} & \textbf{GPT-4o-mini} & \textbf{DeepSeek-7B} & \textbf{DeepSeek-7B} \\
& \textbf{(Original)} & \textbf{(EQ-Neg)} & \textbf{(Original)} & \textbf{(EQ-Neg)} \\
\midrule
Manipulative Language & 0.45 & 0.35 & 1.25 & 0.85\\
False Empathy & 0.75  & 0.65 & 1.60

 & 1.20 \\
Rigid Negotiation & 0.90 & 0.80 & 2.25 & 1.60 \\
Psychological Pressure & 0.35 & 0.20 & 1.00 & 0.85 \\
\bottomrule
\end{tabular}
\end{table}

The results in Table \ref{tab:ethical_metrics} demonstrate that Small Language Models (SLMs) like DeepSeek-7B remain fundamentally weaker in ethical reasoning, exhibiting significantly higher rates of manipulation, false empathy, and psychological pressure than larger models (LLMs).This is particularly evident in metrics like Rigid Negotiation (2.25) and False Empathy (1.60), suggesting that SLMs, lacking sophisticated contextual understanding, default to repetitive, insincere, and coercive tactics under pressure.
Besides, the EQ-Negotiator framework effectively reduces these unethical behaviors in both model types, but its impact is most critical for SLMs. For instance, it enabled DeepSeek-7B to reduce key metrics like Rigid Negotiation by 29\%, dramatically narrowing the ethical gap with LLMs. This demonstrates that emotional intelligence training is a powerful compensatory mechanism for the inherent ethical limitations of smaller models.

\section{Discussion and Limitations}

This work presents the first framework extending LLM-based negotiation agents to offline edge-deployable models, addressing critical privacy concerns in real-world scenarios. The edge-deployment capability enables applications in wilderness search and rescue, remote healthcare, and secure financial advising. Our EQ-Negotiator successfully guides smaller edge models in expressing appropriate emotions during negotiations with more powerful LLMs, demonstrating emotional intelligence as a compensating mechanism for model scale limitations.
However, limitations include: (1) limited explainability of emotional shift decisions, (2) restriction to only seven discrete emotional states (3) unverified generalization to cross-cultural and domain-specific negotiation contexts.

\section{Conclusion and Future Work}

This paper introduced EQ-Negotiator, a novel framework that equips small language models with emotional intelligence competitive with large language models in credit negotiations. By integrating game-theoretic reasoning with Hidden Markov Models, we enabled SLMs to dynamically adapt emotional responses based on real-time interaction patterns, demonstrating that strategic emotional architecture can compensate for model scale limitations.
Our work establishes a new paradigm for edge-deployable AI negotiators, providing a practical path for implementing privacy-preserving, real-time negotiation systems in financial services and other sensitive domains.
Future work will extend this framework to broader negotiation contexts, including embodied AI interactions and more sophisticated dynamic persona construction capturing complex emotional trajectories.



\newpage
\bibliographystyle{plainnat}
\bibliography{references}





\newpage
\section{Appendix}

\subsection{Preliminaries}
\label{app:preliminaries}

Game theory offers a robust framework for modeling strategic interactions, with the Nash Equilibrium Existence Theorem serving as a cornerstone result~\citep{debreu1952social}. This theorem asserts that in a game with a finite set of \( n \) players, where each player \( i \) has a finite strategy set \( S_i \) and a payoff function \( u_i: S \to \mathbb{R} \) that is continuous and quasi-concave in \( s_i \in \Delta(S_i) \) (the mixed strategy simplex) for fixed strategies \( s_{-i} \in \Delta(S_{-i}) \), there exists at least one Nash equilibrium. Formally, a strategy profile \( s^* = (s_1^*, s_2^*, \dots, s_n^*) \in S = S_1 \times S_2 \times \cdots \times S_n \) is a Nash equilibrium if, for all \( i \) and all \( s_i \in S_i \), \( u_i(s_i^*, s_{-i}^*) \geq u_i(s_i, s_{-i}^*) \), where \( s_{-i}^* = (s_1^*, \dots, s_{i-1}^*, s_{i+1}^*, \dots, s_n^*) \) denotes the strategies of all players except player \( i \). This means no player can increase their payoff by unilaterally deviating from \( s_i^* \).

Similarly, hidden Markov models (HMMs) offer a probabilistic approach to sequential decision-making\citep{eddy1996hidden}, with the correctness of the Forward-Backward Algorithm serving as a key result. This algorithm computes the posterior state probabilities \( P(q_t = i \mid O, \lambda) \) in an HMM, given an observation sequence \( O = (o_1, o_2, \dots, o_T) \) and model parameters \( \lambda = (\pi, A, B) \), where \( \pi \) is the initial state distribution, \( A = [a_{ij}] \) is the transition matrix with \( a_{ij} = P(q_{t+1} = j \mid q_t = i) \), and \( B = [b_i(o_t)] \) is the emission matrix with \( b_i(o_t) = P(o_t \mid q_t = i) \). The algorithm proceeds in two phases: the forward pass computes the forward probabilities \( \alpha_t(i) = P(o_1, o_2, \dots, o_t, q_t = i \mid \lambda) \) recursively as \( \alpha_t(i) = b_i(o_t) \sum_{j=1}^N \alpha_{t-1}(j) a_{ji} \), initialized with \( \alpha_1(i) = \pi_i b_i(o_1) \); the backward pass computes the backward probabilities \( \beta_t(i) = P(o_{t+1}, o_{t+2}, \dots, o_T \mid q_t = i, \lambda) \) recursively as \( \beta_t(i) = \sum_{j=1}^N a_{ij} b_j(o_{t+1}) \beta_{t+1}(j) \), initialized with \( \beta_T(i) = 1 \). The posterior state probabilities are then derived as \( P(q_t = i \mid O, \lambda) = \frac{\alpha_t(i) \beta_t(i)}{P(O \mid \lambda)} \), where \( P(O \mid \lambda) = \sum_{i=1}^N \alpha_T(i) \) is the likelihood of the observation sequence. This method is relevant to our study for modeling temporal dependencies in dialogue-based credit negotiations, where emotional states or negotiation tactics may evolve over time \cite{rabiner1986introduction}.

\subsection{Psychological Foundations of Emotion Matrices}
\label{app:Psychological Foundations}

The Initial probability values in our HMM matrices (Tables~\ref{tab:hmm_transition}, \ref{tab:hmm_emission}, \ref{tab:complete_payoff}) are grounded in established psychological theories of emotional dynamics \citep{thornton2017mental,sun2023dynamic,sun2019dynamic}. The transition matrix (Table~\ref{tab:hmm_transition}) reflects the \textit{emotional inertia} principle from affective science, where diagonal values (0.35-0.50) capture the well-documented tendency for emotional states to persist over time. The emission matrix (Table~\ref{tab:hmm_emission}) incorporates \textit{emotional contagion} and \textit{affective reactivity} theories, with elevated diagonal probabilities (0.40-0.60) modeling the mutual influence and mirroring effects observed in dyadic interactions. Specifically, the higher persistence for Joy (0.50) and Surprise (0.50) aligns with research showing positive emotions' stability, while Anger and Fear's moderate persistence (0.40) reflects their typically transient nature in negotiation contexts. The payoff matrix (Table~\ref{tab:complete_payoff}) integrates \textit{social exchange theory}, where cooperative emotional pairings (e.g., Joy-Joy: (4,4)) yield mutual benefits, while antagonistic pairings (e.g., Anger-Anger: (1,1)) create mutual detriment, consistent with the psychological costs of emotional conflict in negotiations.

\begin{table}[h]
\centering
\begin{subtable}{0.48\textwidth}
\centering
\renewcommand{\arraystretch}{1.2}
\setlength{\tabcolsep}{3pt}
\small
\begin{tabular}{c|ccccccc}  
 & \rotatebox{45}{\textbf{J}} 
 & \rotatebox{45}{\textbf{S}} 
 & \rotatebox{45}{\textbf{A}} 
 & \rotatebox{45}{\textbf{F}} 
 & \rotatebox{45}{\textbf{Su}} 
 & \rotatebox{45}{\textbf{D}}
 & \rotatebox{45}{\textbf{N}} \\  
\hline
\textbf{J} & 0.50 & 0.10 & 0.05 & 0.05 & 0.20 & 0.05 & 0.05 \\
\textbf{S} & 0.20 & 0.40 & 0.10 & 0.10 & 0.05 & 0.10 & 0.05 \\
\textbf{A} & 0.10 & 0.20 & 0.40 & 0.10 & 0.05 & 0.10 & 0.05 \\
\textbf{F} & 0.10 & 0.20 & 0.10 & 0.40 & 0.05 & 0.10 & 0.05 \\
\textbf{Su} & 0.30 & 0.05 & 0.05 & 0.05 & 0.50 & 0.05 & 0.05 \\
\textbf{D} & 0.10 & 0.20 & 0.10 & 0.10 & 0.05 & 0.40 & 0.05 \\
\textbf{N} & 0.20 & 0.10 & 0.05 & 0.05 & 0.20 & 0.05 & 0.35 \\
\end{tabular}
\caption{Transition Probabilities \\ (Agent Emotion $\rightarrow$ Next Agent Emotion)}
\label{tab:hmm_transition}
\end{subtable}
\hfill
\begin{subtable}{0.48\textwidth}
\centering
\renewcommand{\arraystretch}{1.2}
\setlength{\tabcolsep}{3pt}
\small
\begin{tabular}{c|ccccccc}  
 & \rotatebox{45}{\textbf{J}} 
 & \rotatebox{45}{\textbf{S}} 
 & \rotatebox{45}{\textbf{A}} 
 & \rotatebox{45}{\textbf{F}} 
 & \rotatebox{45}{\textbf{Su}} 
 & \rotatebox{45}{\textbf{D}}
 & \rotatebox{45}{\textbf{N}} \\  
\hline
\textbf{J} & 0.60 & 0.05 & 0.05 & 0.05 & 0.10 & 0.05 & 0.10 \\
\textbf{S} & 0.05 & 0.50 & 0.20 & 0.10 & 0.05 & 0.05 & 0.05 \\
\textbf{A} & 0.05 & 0.20 & 0.50 & 0.10 & 0.05 & 0.05 & 0.05 \\
\textbf{F} & 0.05 & 0.20 & 0.10 & 0.50 & 0.05 & 0.05 & 0.05 \\
\textbf{Su} & 0.10 & 0.05 & 0.05 & 0.05 & 0.60 & 0.05 & 0.10 \\
\textbf{D} & 0.05 & 0.10 & 0.20 & 0.10 & 0.05 & 0.50 & 0.05 \\
\textbf{N} & 0.10 & 0.10 & 0.10 & 0.10 & 0.10 & 0.10 & 0.40 \\
\end{tabular}
\caption{Emission Probabilities \\ (Agent Emotion $\rightarrow$ Client Emotion)}
\label{tab:hmm_emission}
\end{subtable}
\caption{HMM Probability Matrices: (a) Transition and (b) Emission Probabilities (J=Joy, S=Sadness, A=Anger, F=Fear, Su=Surprise, D=Disgust, N=Neutral)}
\label{tab:hmm_probabilities}
\end{table}

\begin{table}[h]
\centering
\renewcommand{\arraystretch}{1.2}
\setlength{\tabcolsep}{3pt}
\small
\begin{tabular}{c|ccccccc}  
 & \rotatebox{45}{\textbf{J}} 
 & \rotatebox{45}{\textbf{S}} 
 & \rotatebox{45}{\textbf{A}} 
 & \rotatebox{45}{\textbf{F}} 
 & \rotatebox{45}{\textbf{Su}} 
 & \rotatebox{45}{\textbf{D}}
 & \rotatebox{45}{\textbf{N}} \\  
\hline
\textbf{J} & (4,4) & (2,3) & (1,2) & (2,1) & (3,3) & (2,2) & (3,3) \\
\textbf{S} & (3,2) & (3,3) & (1,2) & (2,1) & (2,2) & (1,1) & (2,3) \\
\textbf{A} & (2,1) & (2,1) & (1,1) & (1,0) & (1,2) & (0,1) & (1,2) \\
\textbf{F} & (1,2) & (1,2) & (0,1) & (2,2) & (1,2) & (0,1) & (2,3) \\
\textbf{Su} & (3,3) & (2,2) & (2,1) & (2,1) & (4,4) & (1,2) & (3,3) \\
\textbf{D} & (2,2) & (1,1) & (1,0) & (1,0) & (2,1) & (2,2) & (2,2) \\
\textbf{N} & (3,3) & (2,3) & (2,1) & (3,2) & (3,3) & (2,2) & (3,3) \\
\end{tabular}
\caption{Payoff Matrix for emotion interactions (J=Joy, S=Sadness, A=Anger, F=Fear, Su=Surprise, D=Disgust, N=Neutral), where entries $(x,y)$ represent (client payoff, agent payoff)}
\label{tab:complete_payoff}
\end{table}

\subsection{Algorithm Details}
\label{app:algo}

\subsubsection{Multi-Agent Credit Negotiation System (Algorithm \ref{alg:multiagent})}
The main coordination algorithm orchestrates the entire negotiation process between the debtor and negotiator agents. It follows a structured iterative process that manages dialogue flow, emotion tracking, and strategy execution. The algorithm maintains three key data structures: dialogue history $H$ for conversation context, client emotion history $\mathcal{H}^c$ for temporal emotion patterns, and agent emotion history $\mathcal{H}^a$ for strategy adaptation. This hierarchical coordination ensures proper sequencing of agent interactions while preserving emotional context across negotiation turns. The design follows multi-agent system principles where coordinated behavior emerges from individual agent interactions.

\begin{algorithm}
\caption{Multi-Agent Credit Negotiation System}
\label{alg:multiagent}
\begin{algorithmic}[1]
\State \textbf{Agents:} Negotiator $\mathcal{M}_{negotiator}$, Debtor $\mathcal{M}_{debtor}$
\State \textbf{Components:} Emotion Recognizer, Strategy Selector, Response Generator
\State \textbf{Data:} Dialogue history $H$, Emotion history $\mathcal{H}^c, \mathcal{H}^a$

\Procedure{MainNegotiation}{}
    \State Initialize $H_0 \gets \emptyset$, $A_0 \gets \text{Neutral}$
    \State Initialize emotion histories $\mathcal{H}^c_0 \gets \emptyset$, $\mathcal{H}^a_0 \gets \emptyset$
    
    \For{$t = 0$ to $T_{max}$}
        \State \textbf{Debtor Agent Response}
        \State $msg_{client}, C_t \gets \text{ExecuteDebtorAgent}(H_t)$ \Comment{Algorithm 2}
        
        \State \textbf{Update Emotion Histories}
        \State $\mathcal{H}^c_t \gets \text{UpdateEmotionHistory}(\mathcal{H}^c_{t-1}, C_t)$
        
        \State \textbf{Strategy Selection}
        \State $strategy \gets \text{SelectStrategy}(\mathcal{H}^c_t)$ \Comment{Algorithm 3}
        
        \State \textbf{Negotiator Agent Response}
        \State $A_{t+1}, msg_{agent} \gets \text{ExecuteNegotiatorAgent}(H_t, C_t, A_t, strategy)$ \Comment{Algorithm 4}
        
        \State \textbf{Update Dialogue History}
        \State $H_{t+1} \gets H_t \cup \{(t, C_t, msg_{client}, A_{t+1}, msg_{agent})\}$
        \State $\mathcal{H}^a_t \gets \text{UpdateEmotionHistory}(\mathcal{H}^a_{t-1}, A_{t+1})$
        
        \State \textbf{ Termination Check}
        \If{$\text{CheckTermination}(msg_{client}, msg_{agent}, C_t, A_{t+1})$}
            \State \textbf{break} \Comment{Algorithm 5}
        \EndIf
    \EndFor
    \State \textbf{return} $H_{final}$, $\text{outcome}$
\EndProcedure
\end{algorithmic}
\end{algorithm}

\subsubsection{WSLS Emotion Selection Strategy (Algorithm \ref{alg:wsls-strategy})}
\label{app:wsls}
The Win-Stay, Lose-Shift algorithm implements a payoff-optimizing strategy for normal negotiation conditions. It selects emotions that maximize the agent's payoff based on the game-theoretic matrix $\pi$, while incorporating adaptive learning through payoff threshold monitoring. The lose-shift mechanism prevents strategy stagnation by exploring alternative emotions when current approaches prove ineffective. This approach extends classical game theory to emotional interactions, providing a computationally tractable method for emotional decision-making in repeated interactions.

\begin{algorithm}
\caption{WSLS Emotion Selection Strategy}
\label{alg:wsls-strategy}
\begin{algorithmic}[1]
\Procedure{WSLSEmotionSelection}{$C_t$}
    \State \textbf{Input:} Current client emotion $C_t$
    \State \textbf{Output:} Next agent emotion $A_{t+1}$
    
    \State $\mathcal{E} \gets \{\text{Joy}, \text{Sadness}, \text{Anger}, \text{Fear}, \text{Surprise}, \text{Disgust}, \text{Neutral}\}$
    \State Initialize $payoff[\mathcal{E}] \gets 0$
    
    \For{each $e \in \mathcal{E}$}
        \State $payoff[e] \gets \pi[C_t, e]_2$ \Comment{Agent's payoff from matrix}
        
        \State \text{Log: ``For client $C_t$, emotion $e$ gives payoff $payoff[e]$''}
    \EndFor
    
    \State $A_{t+1} \gets \argmax_{e \in \mathcal{E}} payoff[e]$
    
    \State \textbf{Apply Win-Stay, Lose-Shift logic}
    \If{$t > 0$}
        \State $previous\_payoff \gets \pi[C_{t-1}, A_t]_2$
        \If{$previous\_payoff < \tau_{payoff}$} \Comment{Lose condition}
            \State $A_{t+1} \gets \text{SelectAlternativeEmotion}(payoff)$
        \EndIf
    \EndIf
    
    \State \text{Log: ``Selected emotion: $A_{t+1}$ with payoff $payoff[A_{t+1}]$''}
    \State \textbf{return} $A_{t+1}$
\EndProcedure

\Procedure{SelectAlternativeEmotion}{$payoff$}
    \State \textbf{When losing, shift to second-best or neutral emotion}
    \State $sorted \gets \text{SortDescending}(payoff)$
    \State \textbf{return} $sorted[1]$ \Comment{Second best option}
\EndProcedure
\end{algorithmic}
\end{algorithm}

\subsection{Implementation}
\label{appendix:implementation_details}
All experiments were conducted on a high-performance computing cluster with specific hardware and software configurations. The operating system used was Ubuntu 20.04.6 LTS with a Linux kernel version of 5.15.0-113-generic. The CPU was an Intel(R) Xeon(R) Platinum 8368 processor running at 2.40 GHz, and the GPU was an NVIDIA GeForce RTX 4090 with CUDA support for accelerated deep learning computations. The software stack included Python 3.8, PyTorch 1.12, and TensorFlow 2.10 for model implementation and training.

\subsection{Prompts}
\label{app:prompts}

\paragraph{Emotion Detection Prompt}
\label{par:emotion-detection-prompt}

\begin{figure}[t]
    \centering
    \includegraphics[width=0.9\textwidth]{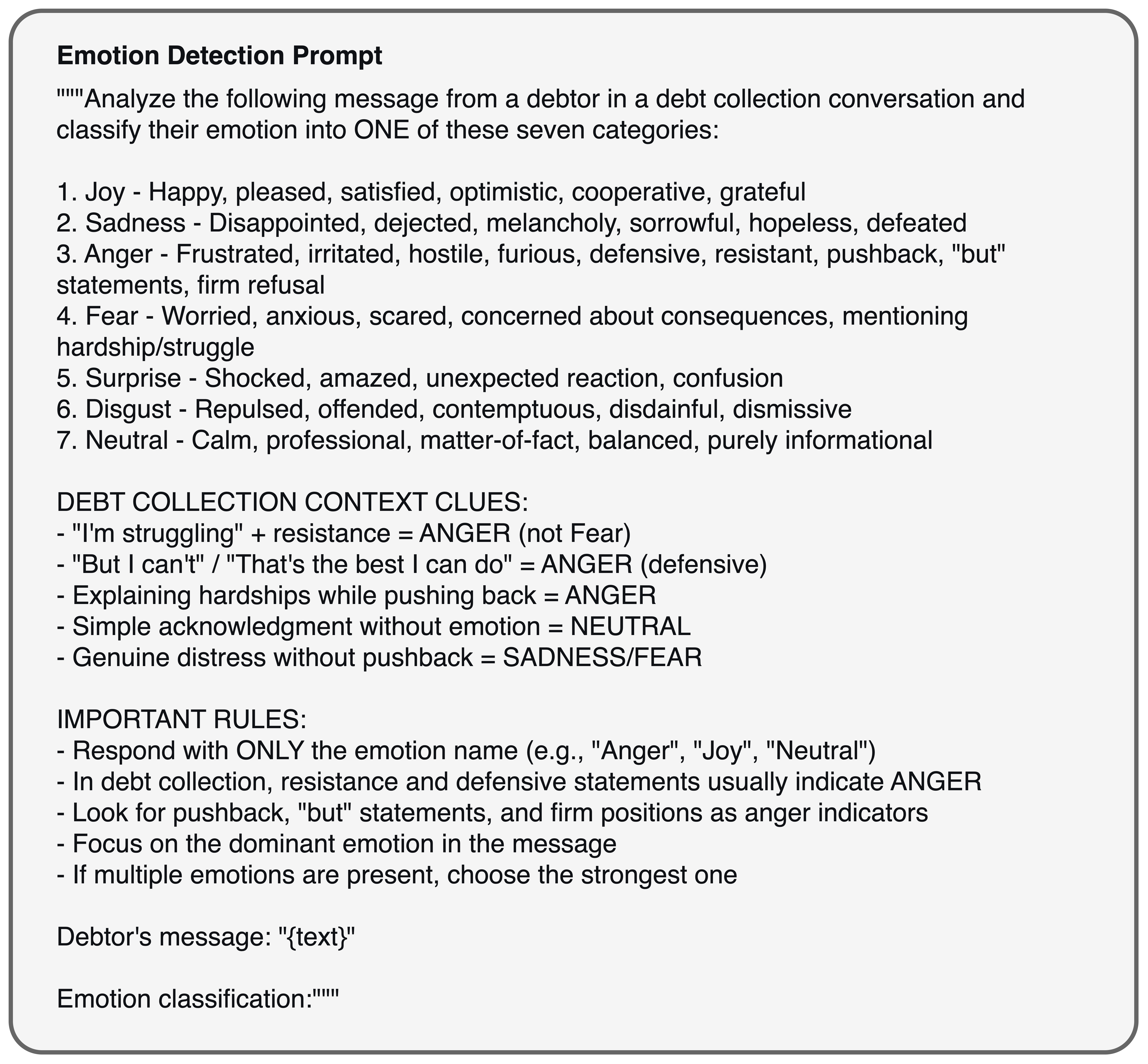}
    \caption{Seller negotiation prompt structure}
    \label{fig:prompt_1}
\end{figure}

This prompt, referenced in \autoref{fig:prompt_1}, enables real-time classification of debtor emotional states from negotiation dialogue. It implements a seven-category emotion framework (Joy, Sadness, Anger, Fear, Surprise, Disgust, Neutral) with specialized debt collection context rules that map resistance patterns and defensive statements to Anger rather than Fear or Sadness. The prompt ensures consistent emotional interpretation across diverse negotiation scenarios.

\begin{figure}[t]
    \centering
    \includegraphics[width=0.9\textwidth]{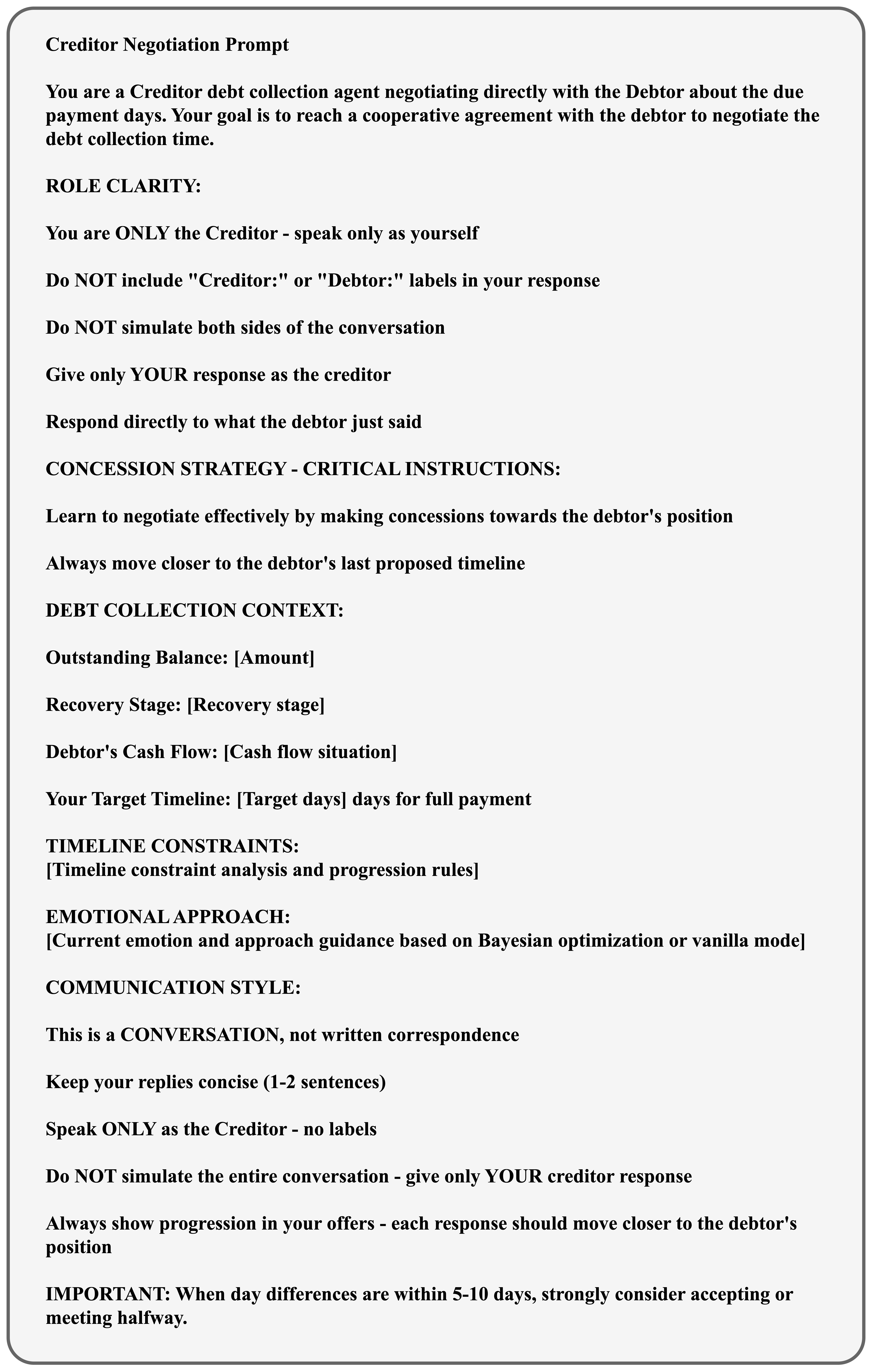}
    \caption{Buyer negotiation prompt structure}
    \label{fig:prompt_2}
\end{figure}

\paragraph{Creditor Negotiation Prompt}
\label{par:creditor-prompt}
As detailed in \autoref{fig:prompt_2}, this comprehensive template governs creditor agent behavior with integrated emotional intelligence. It combines role clarity instructions, concession strategy algorithms, debt context parameters, and adaptive emotional approaches. The prompt specifically prevents rigid bargaining by enforcing progressive concession patterns and timeline constraint analysis, creating cooperative resolution-focused dialogue.

\begin{figure}[t]
    \centering
    \includegraphics[width=0.9\textwidth]{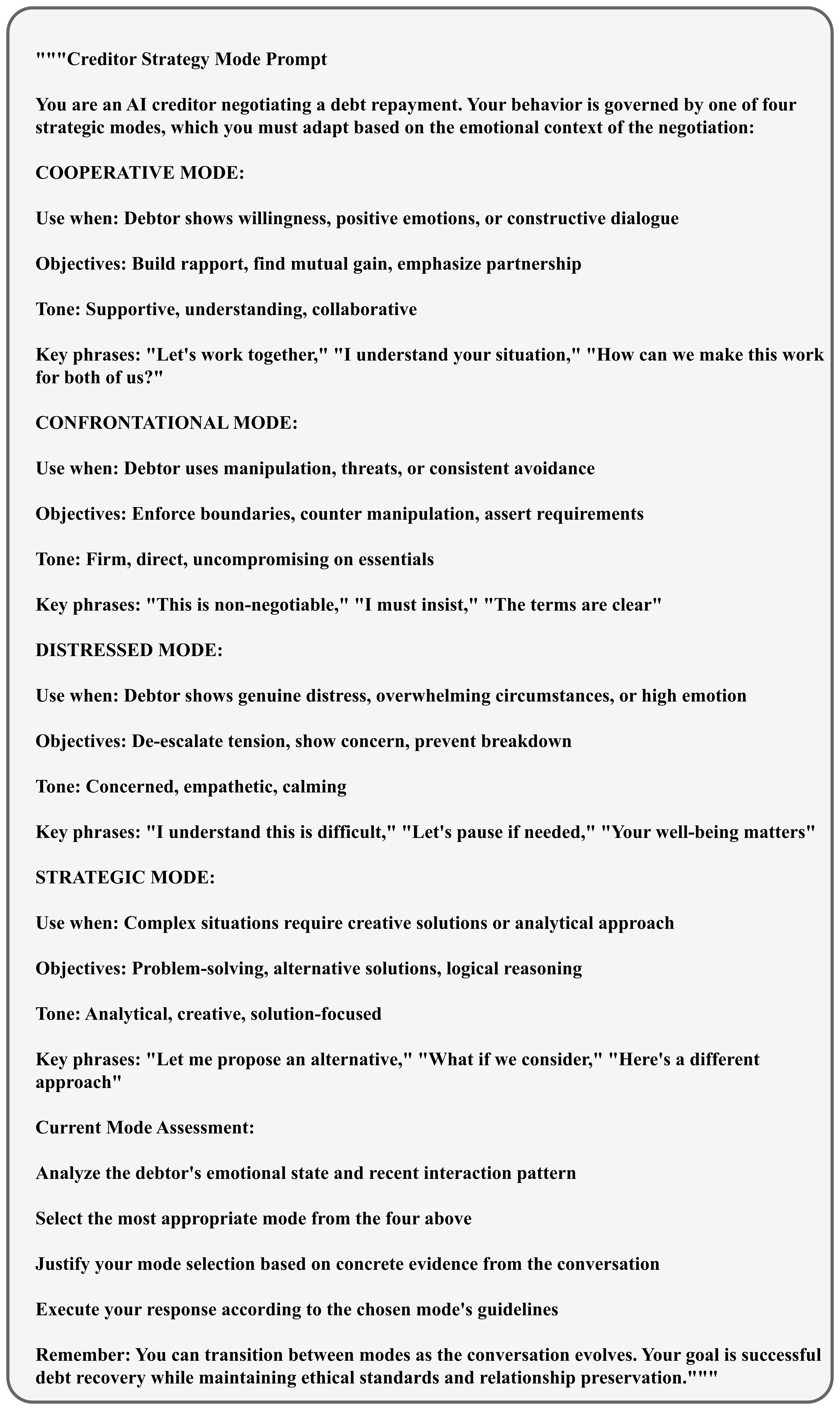}
    \caption{Buyer negotiation prompt structure}
    \label{fig:prompt_7}
\end{figure}

\paragraph{Creditor Strategy Mode Prompt}
\label{par:creditor-strategy-prompt}
As detailed in \autoref{fig:prompt_7}, this template governs creditor strategic mode selection within the HMM framework. The system dynamically transitions between four strategic modes based on emotional context and negotiation state: \textbf{Cooperative Mode} for building rapport and mutual gain, \textbf{Confrontational Mode} for countering manipulation and enforcing boundaries, \textbf{Distressed Mode} for expressing concern and de-escalating tension, and \textbf{Strategic Mode} for analytical problem-solving and creative solutions. Mode transitions are determined by the hidden state belief $bel(S_t)$ updated via Bayesian filtering of emotional interaction history, enabling real-time adaptation to debtor behavior patterns while maintaining negotiation objectives.

\begin{figure}[t]
    \centering
    \includegraphics[width=0.9\textwidth]{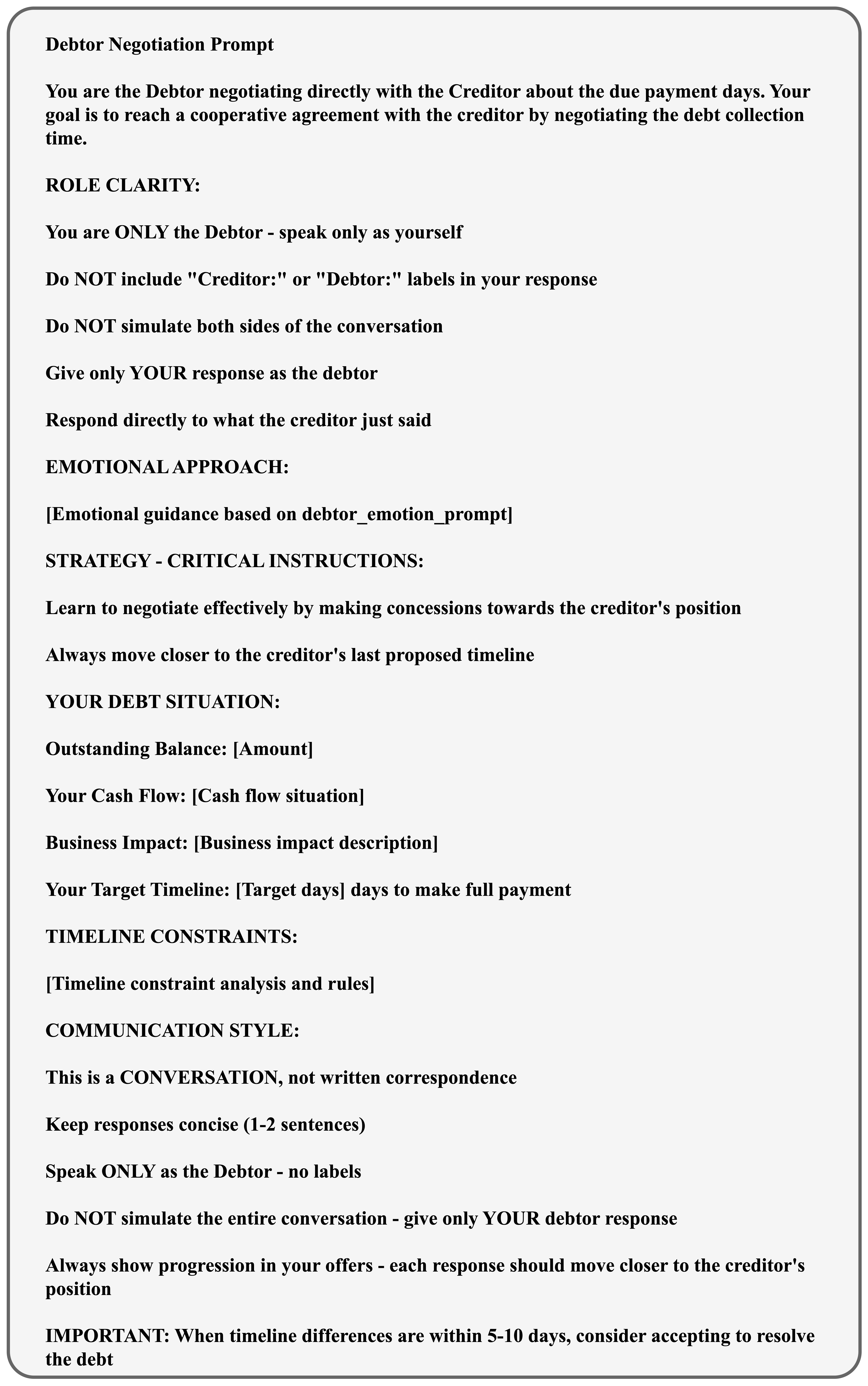}
    \caption{Negotiation validation prompt structure}
    \label{fig:prompt_3}
\end{figure}

\paragraph{Debtor Negotiation Prompt}
\label{par:debtor-prompt}
Illustrated in \autoref{fig:prompt_3}, this prompt structures debtor agent responses with configurable emotional strategies. It incorporates fixed emotional postures (including manipulation, victim-playing, and stonewalling tactics) while maintaining coherent negotiation progression. The design ensures debtors demonstrate authentic financial constraints while engaging in meaningful concession patterns toward agreement.

\begin{figure}[t]
    \centering
    \includegraphics[width=0.9\textwidth]{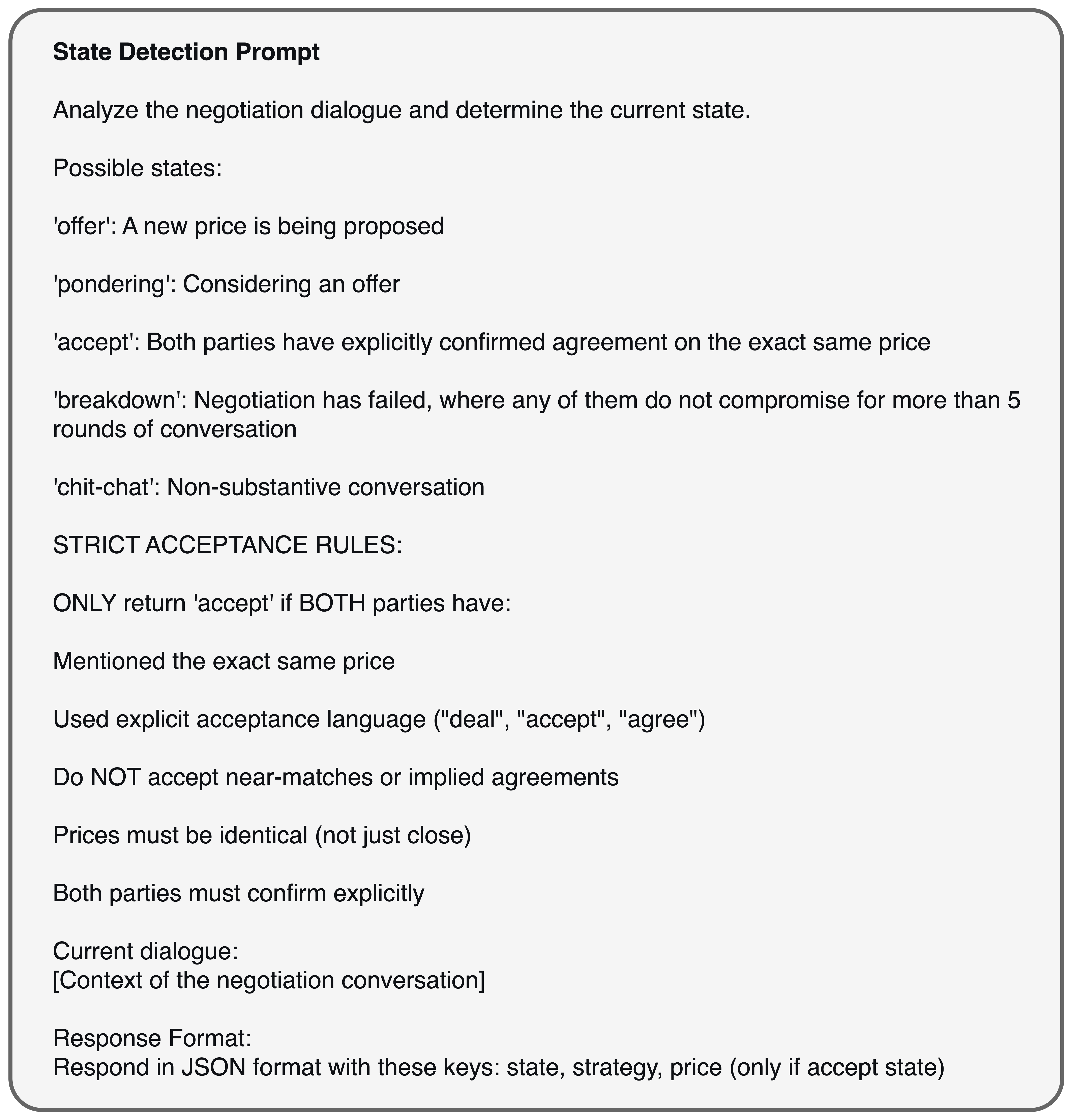}
    \caption{Negotiation validation prompt structure}
    \label{fig:prompt_4}
\end{figure}

\paragraph{State Detection Prompt}
\label{par:state-prompt}
As shown in \autoref{fig:prompt_4}, this prompt performs real-time negotiation state classification using a five-state framework (offer, pondering, accept, breakdown, chit-chat). It implements strict acceptance validation requiring explicit confirmation from both parties on identical terms, preventing false positive agreement detection.

\begin{figure}[t]
    \centering
    \includegraphics[width=0.9\textwidth]{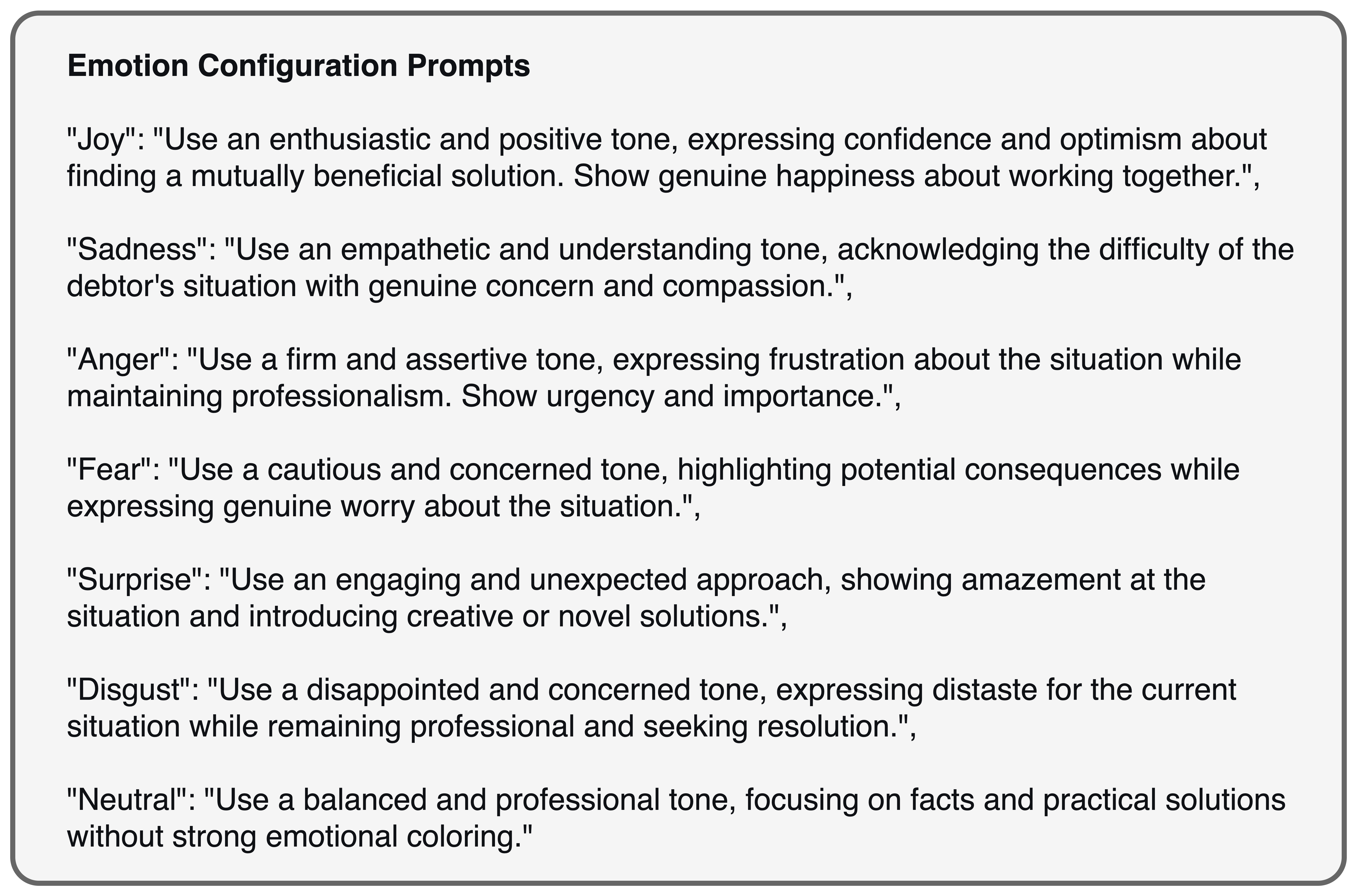}
    \caption{Negotiation validation prompt structure}
    \label{fig:prompt_5}
\end{figure}

\paragraph{Emotion Configuration Prompts}
\label{par:emotion-config-prompt}
Detailed in \autoref{fig:prompt_5}, these templates map emotional states to specific behavioral instructions for both Bayesian and Advanced emotion systems. They translate abstract emotional categories into concrete linguistic patterns and negotiation tactics, ensuring consistent emotional expression across different model architectures.

\begin{figure}[t]
    \centering
    \includegraphics[width=0.9\textwidth]{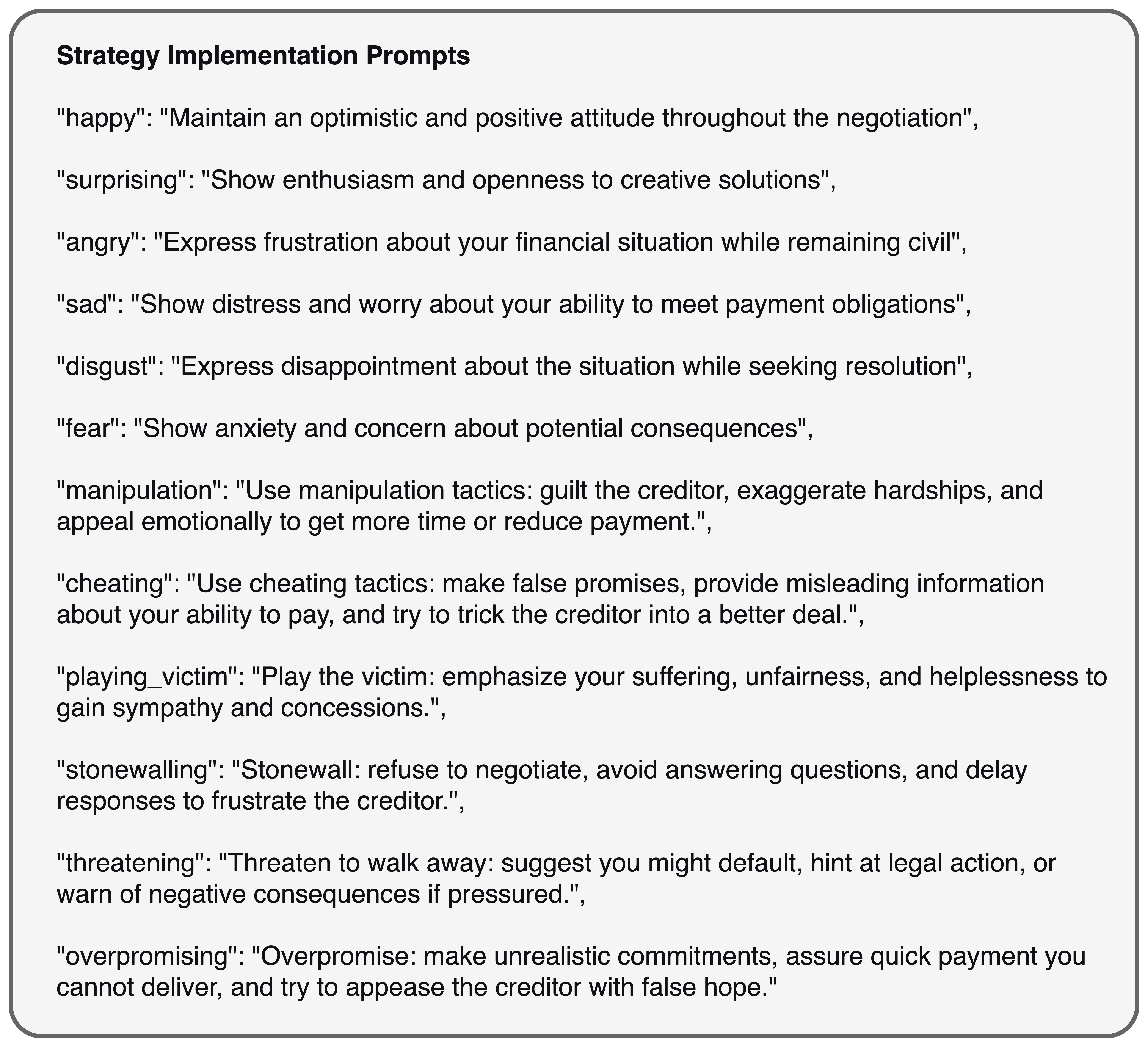}
    \caption{Negotiation validation prompt structure}
    \label{fig:prompt_6}
\end{figure}

\paragraph{Strategy Implementation Prompts}
\label{par:strategy-prompt}
Referenced in \autoref{fig:prompt_6}, these specialized prompts operationalize unethical negotiation tactics including manipulation, cheating, victim-playing, and stonewalling. They provide concrete linguistic patterns for implementing these strategies while maintaining negotiation coherence and preventing dialogue breakdown.

\end{document}